# K-vec: A New Approach for Aligning Parallel Texts

Pascale Fung
Computer Science Department
Columbia University
New York, NY 10027 USA
fung@cs.columbia.edu

Kenneth Ward Church
AT&T Bell Laboratories
600 Mountain Ave.
Murray Hill, NJ 07974 USA
kwc@research.att.com

**Abstract**

Various methods have been proposed for aligning texts in two or more languages such as the Canadian Parliamentary Debates (Hansards). Some of these methods generate a bilingual lexicon as a by-product. We present an alternative alignment strategy which we call K-vec, that starts by estimating the lexicon. For example, it discovers that the English word *fisheries* is similar to the French *pêches* by noting that the distribution of *fisheries* in the English text is similar to the distribution of *pêches* in the French. K-vec does not depend on sentence boundaries.

## 1. Motivation

There have been quite a number of recent papers on parallel text: Brown *et al* (1990, 1991, 1993), Chen (1993), Church (1993), Church *et al* (1993), Dagan *et al* (1993), Gale and Church (1991, 1993), Isabelle (1992), Kay and Rösenschein (1993), Klavans and Tzoukermann (1990), Kupiec (1993), Matsumoto (1991), Ogden and Gonzales (1993), Shemtov (1993), Simard *et al* (1992), Warwick-Armstrong and Russell (1990), Wu (to appear). Most of this work has been focused on European language pairs, especially English-French. It remains an open question how well these methods might generalize to other language pairs, especially pairs such as English-Japanese and English-Chinese.

In previous work (Church *et al*, 1993), we have reported some preliminary success in aligning the English and Japanese versions of the *AWK* manual (Aho, Kernighan, Weinberger (1980)), using *char_align* (Church, 1993), a method that looks for character sequences that are the same in both the source and target. The *char_align* method was designed for European language pairs, where cognates often share character sequences, e.g., *government* and *gouvernement*. In general, this approach doesn't work between languages such as English and Japanese which are written in different alphabets. The *AWK* manual happens to contain a large number of examples and technical words that are the same in the English source and target Japanese.

It remains an open question how we might be able to align a broader class of texts, especially those that are written in different character sets and share relatively few character sequences. The K-vec method attempts to address this question.

## 2. The K-vec Algorithm

K-vec starts by estimating the lexicon. Consider the example: *fisheries* → *pêches*. The K-vec algorithm will discover this fact by noting that the distribution of *fisheries* in the English text is similar to the distribution of *pêches* in the French.

The concordances for *fisheries* and *pêches* are shown in Tables 1 and 2 (at the end of this paper).[1]

---

1. These tables were computed from a small fragment of the Canadian Hansards that has been used in a number of other studies: Church (1993) and Simard *et al* (1992). The English text has 165,160 words and the French text has 185,615 words.

There are 19 instances of *fisheries* and 21 instances of *pêches*. The numbers along the left hand edge show where the concordances were found in the texts. We want to know whether the distribution of numbers in Table 1 is similar to those in Table 2, and if so, we will suspect that *fisheries* and *pêches* are translations of one another. A quick look at the two tables suggests that the two distributions are probably very similar, though not quite identical.[2]

We use a simple representation of the distribution of *fisheries* and *pêches*. The English text and the French text were each split into K pieces. Then we determine whether or not the word in question appears in each of the K pieces. Thus, we denote the distribution of *fisheries* in the English text with a K-dimensional binary vector, $V_f$, and similarly, we denote the distribution of *pêches* in the French text with a K-dimensional binary vector, $V_p$. The $i^{th}$ bit of $V_f$ indicates whether or not *Fisheries* occurs in the $i^{th}$ piece of the English text, and similarly, the $i^{th}$ bit of $V_p$ indicates whether or not *pêches* occurs in the $i^{th}$ piece of the French text.

If we take K be 10, the first three instances of *fisheries* in Table 1 fall into piece 2, and the remaining 16 fall into piece 8. Similarly, the first 4 instances of *pêches* in Table 2 fall into piece 2, and the remaining 17 fall into piece 8. Thus,

$$V_f = V_p = <0,0,1,0,0,0,0,0,1,0>$$

Now, we want to know if $V_f$ is similar to $V_p$, and if we find that it is, then we will suspect that *fisheries* → *pêches*. In this example, of course, the vectors are identical, so practically any reasonable similarity statistic ought to produce the desired result.

### 3. *fisheries* is not the translation of *lections*

Before describing how we estimate the similarity of $V_f$ and $V_p$, let us see what would happen if we tried to compare *fisheries* with a completely unrelated word, eg., *lections*. (This word should be the translation of *elections*, not *fisheries*.)

---

2. At most, *fisheries* can account for only 19 instances of *pêches*, leaving at least 2 instances of *pêches* unexplained.

As can be seen in the concordances in Table 3, for K=10, the vector is <1, 1, 0, 1, 1, 0, 1, 0, 0, 0>. By almost any measure of similarity one could imagine, this vector will be found to be quite different from the one for *fisheries*, and therefore, we will correctly discover that *fisheries* is not the translation of *lections*.

To make this argument a little more precise, it might help to compare the contingency matrices in Tables 5 and 6. The contingency matrices show: (a) the number of pieces where both the English and French word were found, (b) the number of pieces where just the English word was found, (c) the number of pieces where just the French word was found, and (d) the number of peices where neither word was found.

Table 4: A contingency matrix

|         | French |   |
|---------|--------|---|
| English | a      | b |
|         | c      | d |

Table 5: *fisheries* vs. *pêches*

|           | pêches |   |
|-----------|--------|---|
| *fisheries* | 2      | 0 |
|           | 0      | 8 |

Table 6: *fisheries* vs. *lections*

|           | lections |   |
|-----------|----------|---|
| *fisheries* | 0        | 2 |
|           | 4        | 4 |

In general, if the English and French words are good translations of one another, as in Table 5, then *a* should be large, and *b* and *c* should be small. In contrast, if the two words are not good translations of one another, as in Table 6, then *a* should be small, and *b* and *c* should be large.

### 4. Mutual Information

Intuitively, these statements seem to be true, but we need to make them more precise. One could have chosen quite a number of similarity metrics for this purpose. We use mutual information:

$$\log_2 \frac{prob(V_f, V_p)}{prob(V_f)\ prob(V_p)}$$

That is, we want to compare the probability of seeing *fisheries* and *pêches* in the same piece to chance. The probability of seeing the two words in the same piece is simply:

$$prob(V_f, V_p) = \frac{a}{a+b+c+d}$$

The marginal probabilities are:

$$prob(V_f) = \frac{a+b}{a+b+c+d}$$

$$prob(V_p) = \frac{a+c}{a+b+c+d}$$

For *fisheries* → *pêches*, $prob(V_f, V_p) = prob(V_f) = prob(V_p) = 0.2$. Thus, the mutual information is $\log_2 5$ or 2.32 bits, meaning that the joint probability is 5 times more likely than chance. In contrast, for *fisheries* → *lections*, $prob(V_f, V_p) = 0$, $prob(V_f) = 0.5$ and $prob(V_p) = 0.4$. Thus, the mutual information is $\log_2 0$, meaning that the joint is infinitely less likely than chance. We conclude that it is quite likely that *fisheries* and *pêches* are translations of one another, much more so than *fisheries* and *lections*.

### 5. Significance

Unfortunately, mutual information is often unreliable when the counts are small. For example, there are lots of infrequent words. If we pick a pair of these words at random, there is a very large chance that they would receive a large mutual information value by chance. For example, let *e* be an English word that appeared just once and let *f* be a French word that appeared just once. Then, there is a non-trivial chance ($\frac{1}{K}$) that *e* and *f* will appear in the same piece, as shown in Table 7. If this should happen, the mutual information estimate would be very large, i.e., $\log K$, and probably misleading.

Table 7:

|   | f |   |
|---|---|---|
| e | 1 | 0 |
|   | 0 | 9 |

In order to avoid this problem, we use a *t*-score to filter out insignificant mutual information values.

$$t \approx \frac{prob(V_f, V_p) - prob(V_f)\ prob(V_p)}{\sqrt{\frac{1}{K}\ prob(V_f, V_p)}}$$

Using the numbers in Table 7, $t \approx 1$, which is not significant. (A *t* of 1.65 or more would be significant at the $p > 0.95$ confidence level.)

Similarly, if *e* and *f* appeared in just two pieces each, then there is approximately a $\frac{1}{K^2}$ chance that they would both appear in the same two pieces, and then the mutual information score would be quite high, $\log \frac{K}{2}$, but we probably wouldn't believe it because the *t*-score would be only $\sqrt{2}$. By this definition of significance, we need to see the two words in at least 3 different pieces before the result would be considered significant.

This means, unfortunately, that we would reject *fisheries* → *pêches* because we found them in only two pieces. The problem, of course, is that we don't have enough pieces. When K=10, there simply isn't enough resolution to see what's going on. At K=100, we obtain the contingency matrix shown in Table 8, and the *t*-score is significant ($t \approx 2.1$).

Table 8: K=100

|   | pêches |   |
|---|---|---|
| fisheries | 5 | 1 |
|   | 0 | 94 |

How do we choose K? As we have seen, if we choose too small a K, then the mutual information values will be unreliable. However, we can only increase K up to a point. If we set K to a ridiculously large value, say the size of the English text, then an English word and its translations are likely to fall in slightly different pieces due to random fluctuations and we would miss the signal. For this work, we set K to the square root of the size of the corpus.

K should be thought of as a scale parameter. If we use too low a resolution, then everything turns into a blur and it is hard to see anything. But if we use too high a resolution, then we can miss the signal if it isn't just exactly where we are looking.

Ideally, we would like to apply the K-vec algorithm to all pairs of English and French words, but unfortunately, there are too many such pairs to consider. We therefore limited the search to pairs of words in the frequency range: 3-10. This heuristic makes the search practical, and catches many interesting pairs.[3]

## 6. Results

This algorithm was applied to a fragment of the Canadian Hansards that has been used in a number of other studies: Church (1993) and Simard *et al* (1992). The 30 significant pairs with the largest mutual information values are shown in Table 9. As can be seen, the results provide a quick-and-dirty estimate of a bilingual lexicon. When the pair is not a direct translation, it is often the translation of a collocate, as illustrated by *acheteur* → *Limited* and *Santé* → *Welfare*. (Note that some words in Table 9 are spelled with same way in English and French; this information is not used by the K-vec algorithm).

Using a scatter plot technique developed by Church and Helfman (1993) called *dotplot*, we can visualize the alignment, as illustrated in Figure 1. The source text ($N_x$ bytes) is concatenated to the target text ($N_y$ bytes) to form a single input sequence of $N_x + N_y$ bytes. A dot is placed in position $i,j$ whenever the input token at position $i$ is the same as the input token at position $j$.

The equality constraint is relaxed in Figure 2. A dot is placed in position $i,j$ whenever the input token at position $i$ is highly associated with the input token at position $j$ as determined by the mutual information score of their respective K-vecs. In addition, it shows a detailed, magnified and rotated view of the diagonal line. The alignment program tracks this line with as much precision as possible.

---

3. The low frequency words (frequency less then 3) would have been rejected anyways as insignificant.

Table 9: K-vec results

| | French | English |
|---|---|---|
| 3.2 | Beauce | Beauce |
| 3.2 | Comeau | Comeau |
| 3.2 | 1981 | 1981 |
| 3.0 | Richmond | Richmond |
| 3.0 | Rail | VIA |
| 3.0 | pêches | Fisheries |
| 2.8 | Deans | Deans |
| 2.8 | Prud | Prud |
| 2.8 | Prud | homme |
| 2.7 | acheteur | Limited |
| 2.7 | Communications | Communications |
| 2.7 | MacDonald | MacDonald |
| 2.6 | Mazankowski | Mazankowski |
| 2.5 | croisière | nuclear |
| 2.5 | Santé | Welfare |
| 2.5 | 39 | 39 |
| 2.5 | Johnston | Johnston |
| 2.5 | essais | nuclear |
| 2.5 | Université | University |
| 2.5 | bois | lumber |
| 2.5 | Angus | Angus |
| 2.4 | Angus | VIA |
| 2.4 | Saskatoon | University |
| 2.4 | agriculteurs | farmers |
| 2.4 | inflation | inflation |
| 2.4 | James | James |
| 2.4 | Vanier | Vanier |
| 2.4 | Santé | Health |
| 2.3 | royale | languages |
| 2.3 | grief | grievance |

## 7. Conclusions

The K-vec algorithm generates a quick-and-dirty estimate of a bilingual lexicon. This estimate could be used as a starting point for a more detailed alignment algorithm such as *word_align* (Dagan *et al*, 1993). In this way, we might be able to apply *word_align* to a broader class of language combinations including possibly English-Japanese and English-Chinese. Currently, *word_align* depends on *char_align* (Church, 1993) to generate a starting point, which limits its applicability to European languages since *char_align* was designed for language pairs that share a common alphabet.

## References

Aho, Kernighan, Weinberger (1980) ''The AWK Programming Language,'' Addison-Wesley, Reading, Massachusetts, USA.

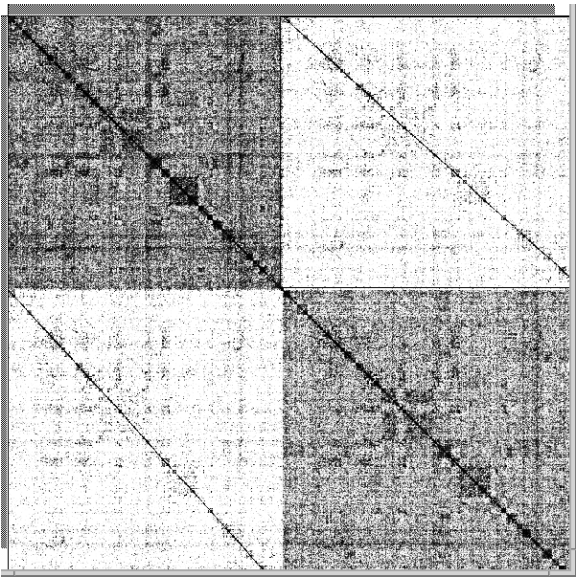

Figure 1: A Dotplot of the Hansards

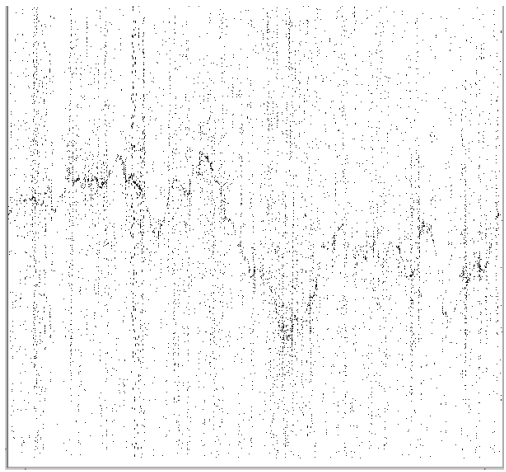

Figure 2: K-vec view of Hansards

Table 1: Concordances for *fisheries*

| | |
|---|---|
| 28312 | Mr . Speaker , my question is for the Minister of Fisheries and Oceans . Allegations have been made |
| 28388 | of the stocks ? Hon . Thomas Siddon ( Minister of Fisheries and Oceans ): Mr . Speaker , I tell the |
| 28440 | calculation on which the provincial Department of Fisheries makes this allegation and I find that it |
| 128630 | private sector is quite weak . Let us turn now to fisheries , an industry which as most important fo |
| 128885 | The fishermen would like to see the Department of Fisheries and Oceans put more effort towards the p |
| 128907 | s in particular . The budget of the Department of Fisheries and Oceans has been reduced to such a le |
| 130887 | ' habitation ' ' from which to base his trade in fisheries and furs . He brought with him the first |
| 132282 | ase just outside of my riding . The Department of Fisheries and Oceans provides employment for many |
| 132629 | and all indications are that the richness of its fisheries resource will enable it to maintain its |
| 132996 | taxpayer . The role of the federal Department of Fisheries and Oceans is central to the concerns of |
| 134026 | is the new Chairman of the Standing Committee on Fisheries and Oceans . I am sure he will bring a w |
| 134186 | ortunity to discuss it with me as a member of the Fisheries Committee . The Hon . Member asked what |
| 134289 | he proposal has been submitted to the Minister of Fisheries and Oceans ( Mr . Siddon ) which I hope |
| 134367 | ch as well as on his selection as Chairman of the Fisheries Committee . I have worked with Mr . Come |
| 134394 | his intense interest and expertise in the area of fisheries . It seems most appropriate , given that |
| 134785 | r from Eastern Canada and the new Chairman of the Fisheries and Oceans Committee . We know that the |
| 134796 | d Oceans Committee . We know that the Minister of Fisheries and Oceans ( Mr . Siddon ) , should we s |
| 134834 | ows the importance of research and development to fisheries and oceans . Is he now ready to tell the |
| 134876 | research and development component in the area of fisheries and oceans at Bedford , in order that th |

Table 2: Concordances for *pêches*

|  |  |
|---|---|
| 31547 | oyez certain que je présenterai mes excuses . Les pêches L ' existence possible d ' un marché noir e |
| 31590 | ésident , ma question s ' adresse au ministre des Pêches et des Océans . On aurait pêché , débarqué |
| 31671 | poissons ? L ' hon . Thomas Siddon ( ministre des Pêches et des Océans ) |
| 31728 | calculs sur lesquels le ministère provincial des pêches fonde ses allégations , et j ' y ai relevé |
| 144855 | ivé est beaucoup plus faible . Parlons un peu des pêches , un secteur très important pour l ' Atlant |
| 145100 | braconnage . Ils voudraient que le ministère des Pêches et des Océans fasse davantage , particulièr |
| 145121 | es stocks de homards . Le budget du ministère des Pêches et des Océans a té amputé de telle sorte qu |
| 148873 | endant l ' hiver , lorsque l ' agriculture et les pêches sont peu près leur point mort , bon nombre |
| 149085 | xtérieur de ma circonscription . Le ministère des Pêches et des Océans assure de l ' emploi bien d ' |
| 149837 | s . Dans le rapport Kirby de 1983 portant sur les pêches de la côte est , on a mal expliqué le systè |
| 149960 | eniers publics . Le rôle du ministère fédéral des Pêches et des Océans se trouve au centre des préoc |
| 151108 | soit le nouveau président du comité permanent des pêches et océans . Je suis sûr que ses vastes conn |
| 151292 | avec moi , en ma qualité de membre du comité des pêches et océans . Le député a demandé quelles per |
| 151398 | is savoir qu ' elle a té proposée au ministre des Pêches et Océans ( M . Siddon ) et j ' espère qu ' |
| 151498 | de son choix au poste de président du comité des pêches . Je travaille avec M . Comeau depuis deux |
| 151521 | et je connais tout l ' intérêt qu ' il porte aux pêches , ainsi que sa compétence cet gard . Cela s |
| 151936 | Est du pays et maintenant président du Comité des pêches et des océans . On sait que le ministre des |
| 151947 | êches et des océans . On sait que le ministre des Pêches et des Océans ( M . Siddon ) a , disons , a |
| 151997 | recherche et du développement dans le domaine des pêches et des océans . Est - il prêt aujourd ' hui |
| 152049 | recherche et du développement dans le domaine des pêches et des océans Bedford afin que ce laboratoi |
| 152168 | s endroits ouˆg ils se trouvent et l ' avenir des pêches dans l ' Est . Le président suppléant ( M . |

Table 3: Concordances for *lections*

|  |  |
|---|---|
| 88 | de prendre la parole aujourd ' hui . Bien que les lections au cours desquelles on nous a lus la tête |
| 207 | ui servent ensemble la Chambre des communes . Les lections qui se sont tenues au début de la deuxièm |
| 12439 | n place les mesures de contrôle suffisantes . Les lections approchaient et les libéraux voulaient me |
| 14999 | reprendre le contenu de son discours lectoral des lections de 1984 . On se rappelle , et tous les Ca |
| 16164 | ertainement et s ' en rappelleront aux prochaines lections de tout ce qui aurait pu leur arriver . L |
| 16386 | n apercevront encore une fois lors des prochaines lections . Des lections , monsieur le Président , |
| 16389 | ncore une fois lors des prochaines lections . Des lections , monsieur le Président , il y en a eu de |
| 16431 | avec eux - mêmes l ' analyse des résultats de ces lections complémentaires , constateront qu ' ils o |
| 17419 | s et ils réagissent . Ils ont réagi aux dernières lections complémentaires et ils réagiront encore a |
| 17427 | émentaires et ils réagiront encore aux prochaines lections . Finalement , monsieur le Président , pa |
| 17438 | t , monsieur le Président , parlant de prochaines lections . . . j ' coutais mon honorable collègue |
| 17461 | M . Layton ) dire tantôt que , antérieurement aux lections de 1984 , les gens de Lachine voulaient u |
| 55169 | étitions . Je suggérerais au Comité permanent des lections , des privilèges et de la procédure d ' t |
| 56641 | ulever cette question au comité des privilèges et lections , car il y a de sérieux doutes sur l ' in |
| 57853 | doivent tre renvoyées au comité des privilèges et lections . J ' ai l ' intention d ' en saisir ce c |
| 59027 | rêt soumettre la question au comité permanent des lections , des privilèges et de la procédure . J ' |
| 67980 | le 16 janvier 1986 . . . M . Hovdebo: Après les lections . M . James: . . . le ministre d ' alors |
| 70161 | tinuer faire ce qu ' ils ont fait depuis quelques lections , c ' est - - dire rejeter le Nouveau par |
| 70456 | que les gens le retiennent jusqu ' aux prochaines lections . De cette façon vous allez tre rejetés d |
| 103132 | donc transmis mon mandat au directeur général des lections , afin de l ' autoriser mettre un nouveau |
| 103186 | , deux députés ont avisé le directeur général des lections d ' une vacance survenue la Chambre ; il |